# Predicting Demand for Air Taxi Urban Aviation Services using Machine Learning Algorithms

**Suchithra Rajendran[a,b,*], Sharan Srinivas[a,b] and Trenton Grimshaw[a]**

[a]*Department of Industrial and Manufacturing Systems Engineering, University of Missouri Columbia, MO 65211, USA*

[b]*Department of Marketing, University of Missouri Columbia, MO 65211, USA*

This is a pre-print version of the manuscript accepted in Journal of Air Transport Management (https://doi.org/10.1016/j.jairtraman.2021.102043)

The article should be cited as follows: Rajendran, Suchithra, Sharan Srinivas, and Trenton Grimshaw. "Predicting demand for air taxi urban aviation services using machine learning algorithms." *Journal of Air Transport Management* 92 (2021): 102043.

# Predicting Demand for Air Taxi Urban Aviation Services using Machine Learning Algorithms

**Suchithra Rajendran[a,b,*], Sharan Srinivas[a,b] and Trenton Grimshaw[a]**

[a]*Department of Industrial and Manufacturing Systems Engineering, University of Missouri Columbia, MO 65211, USA*
[b]*Department of Marketing, University of Missouri Columbia, MO 65211, USA*

## Abstract

This research focuses on predicting the demand for air taxi urban air mobility (UAM) services during different times of the day in various geographic regions of New York City using machine learning algorithms (MLAs). Several ride-related factors (such as month of the year, day of the week and time of the day) and weather-related variables (such as temperature, weather conditions and visibility) are used as predictors for four popular MLAs, namely, logistic regression, artificial neural networks, random forests, and gradient boosting. Experimental results suggest gradient boosting to consistently provide higher prediction performance. Specific locations, certain time periods and weekdays consistently emerged as critical predictors.

## 1. Introduction

In recent years, urban air mobility (UAM), an emerging aviation transportation system that strives to commute passenger or cargo by air using low-altitude aircraft, is being widely investigated (Rajendran and Shulman, 2020; Straubinger et al., 2020). With the radical expansion of the aviation industry (Matsumoto and Domae, 2018), several logistics companies are venturing into this nascent market (Rajendran and Pagel, 2020). Air taxi, a form of on-demand UAM service, is expected to launch in the forthcoming years as a fast, safe and efficient mode of transport for everyday commuters in urban and semi-urban areas (Holden and Goel, 2016). These are compact aircraft that operate using the electric vertical takeoff and landing (eVTOL) technology and have an average capacity of four passengers (Rajendran and Zack, 2019). Thus, the goal of the electric flying taxi services is to not only provide a significantly faster commute, but also to operate in a sustainable manner. Air taxis could also serve as a potential travel option for people commuting outside of regular work hours by easing the process of traveling in and out of the metropolitan cities for entertainment purposes, such as sports events or nightlife activities (Holden et al., 2018). Several companies like Uber, Zephyr Airworks, and Airbus are planning to launch their next form of urban on-demand aviation ride service for citizens in highly condensed cities. They have also discussed their eVTOL design concepts and how the prototypes would be tested; however, city planners are still in the process of taking measures to accommodate this new mode of commute (Garrett-Glaser, 2019). Thus, it is essential to predict the demand for these air taxi services to enable manufacturing companies, investors, and city planners to be equipped with the necessary measures for the launch of this aviation operation.

Two types of physical infrastructure assets are proposed for air taxi operations in the literature, namely, *vertistop and vertiport* (Smith, 1994; Patnoe, 2018; Hasan, 2019). Vertistop is a sophisticated rooftop helipad that typically handles a single air taxi eVTOL at a time. These facilities enable air taxis to land and take off relatively quickly with the sole purpose of embarking/disembarking passengers. On the other hand, vertiport is a larger infrastructure and can accommodate multiple eVTOLs at a given time. Aside from having numerous landing-pad structures for serving customers, other air taxi operations, such as vehicle inspection and maintenance, charging, and docking, are also conducted at vertiports. These stations could either be retrofitted on larger buildings or operate on other infrastructures, such as highway roundabouts and open parking lots (Holden et al., 2018).

The air taxi system operations are expected to be integrated with public transportation or on-demand taxi services, thereby leading to an on-demand door-to-door multi-leg multi-modal transportation (Rajendran and Srinivas, 2020). Specifically, most rides are likely to include three legs or segments, as shown in Figure 1. The first-leg transport consists of customer travel from his/her actual pickup location (such as home and office) to a nearby skyport (i.e., vertiport or vertistop). In the main leg, the customer is flown by air taxi to the destination skyport. Finally, the customer is transported to the actual drop-off location in the third segment. The commute choice for the first and last legs of the trip could include walking, on-demand regular taxi services like Uber or Lyft, subway, and bus.

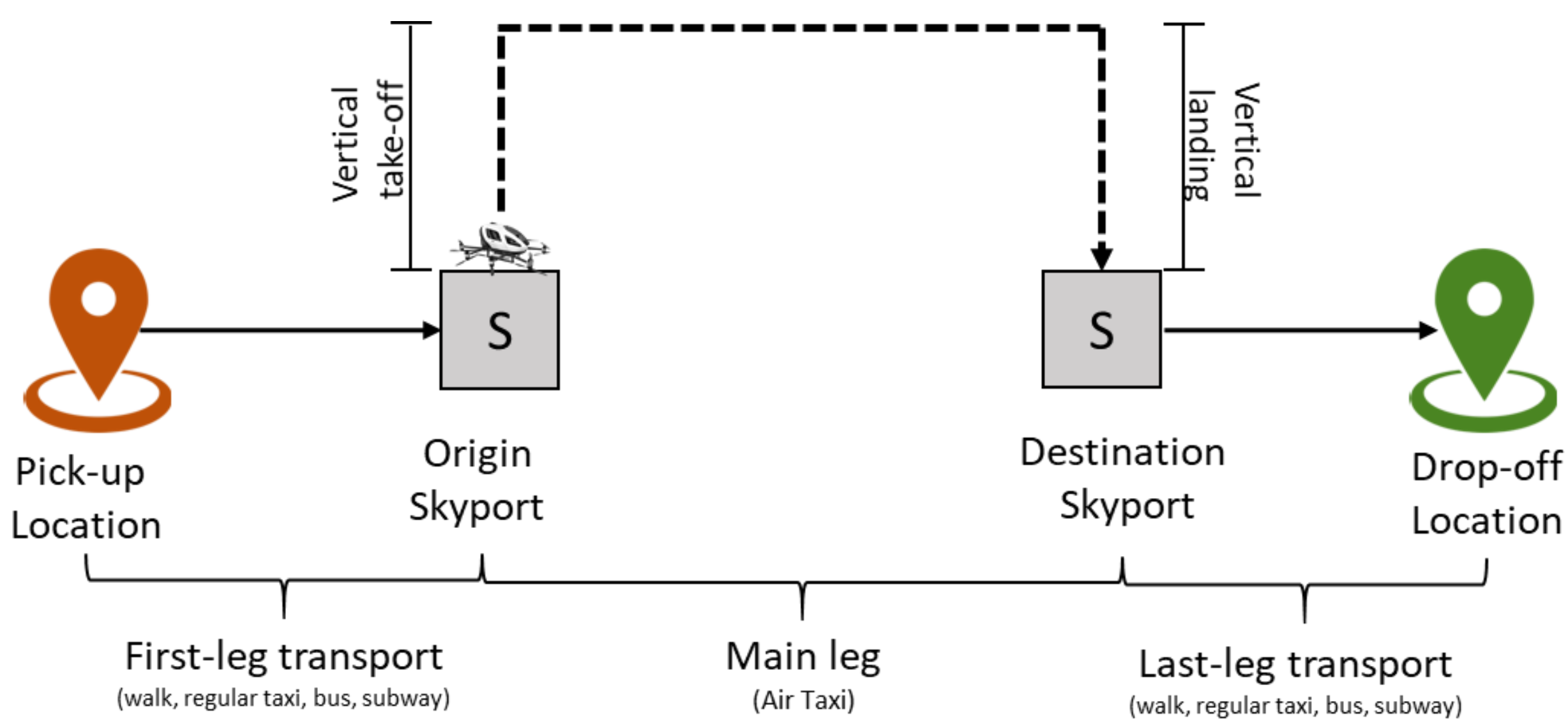

**Figure 1:** Illustration of air taxi operations for door-to-door transportation

Although air taxis have the potential to be a convenient mode of transportation for commuters in metropolitan cities, there are several challenges associated with their implementation (Rajendran and Zack, 2019; Straubinger et al., 2020). For instance, it is essential to establish a system that will make smart real-time dispatching and routing decisions to mitigate customer ride time and costs.

Moreover, the scheduling systems should ensure that flight operations are optimized to increase the demand fulfillment rate. The estimation of passenger demand for air taxi services across time and space is crucial to address the challenges mentioned above and aid several other critical decisions (such as fleet procurement and pricing).

This study aims to develop a data-driven machine-learning-based approach for estimating the spatiotemporal air taxi demand level (low, moderate, or high) using the data established by a prior study on air taxi network design (Rajendran and Zack, 2019). Several ride-related (e.g., pickup and drop-off locations, distance, time of the day, day of the week) and environment-related (e.g., temperature, presence of rain, or snow, visibility) factors are derived from the dataset as well as a commercial weather service provider and used as predictors for machine learning algorithms.
The remainder of the paper is organized as follows. A detailed review of the literature on existing and emerging urban on-demand mobility services are discussed in Section 2. The description of the data used for this study is presented in Section 3. Machine learning algorithms for predicting the air taxi demand is discussed in Section 4. Results are presented in Section 5, while the discussions are detailed in Section 6. Conclusions and future works are given in Section 7.

## 2. Literature Review

In this section, we review the literature pertaining to existing as well as emerging on-demand mobility services and demand prediction for air taxi services.

### 2.1 Existing and Emerging Urban On-Demand Mobility Services

Taxi services have been one of the most popular on-demand urban public transportation methods across the globe for several decades (Zhang et al., 2016; Zhao et al., 2016; Yao et al., 2018). In recent years, the proliferation of technology paired with the rapidly increasing population in metropolitan cities has paved the way for on-demand door-to-door ride-sourcing taxi services such as Uber and Lyft, which offers both individual and shared/pooled ride options to the passengers. These mobility services are reported to easen the traffic congestion in heavily congested urban areas regardless of the "willingness to rideshare" levels (Jeffrey-Wilensky, 2019).
There are a number of studies that provide the framework to start and successfully carry out effective ridesharing services for taxi companies in dense cities like NYC, Atlanta, and Singapore (Chen et al., 2010; Agatz et al., 2011 & 2012; Lin et al., 2012; Ma et al., 2013 & 2015; Ota et al., 2015; Santos and Xavier, 2015; Alonso-Mora et al., 2017; Gurumurthy and Kockelman, 2018; Lokhandwala and Cai, 2018). Several papers have focused on developing a framework for taxi demand prediction (Chang et al., 2010; Li et al., 2017; Liao et al., 2018; Liu et al., 2019). For instance, Moreira-Matias et al. (2013) leveraged streaming data and adopted time series forecasting techniques to predict the passenger demand in 30-minute intervals. However, in recent times, most works have employed machine learning algorithms and used ride-related factors or GPS trace data (e.g., pickup location, drop-off location, pickup time, drop-off time) as predictors for forecasting taxi-passenger demand (Jiang et al., 2019; Zhao et al., 2019; Luo et al., 2020).

The aforementioned prediction models for taxi services provide insights on the spatiotemporal distribution of passenger demand, which, in turn, can be leveraged by models developed for effective dispatching and routing of the vehicle fleet. For example, Alonso-Mora et al. (2018) developed an algorithm based on constrained optimization for real-time matching and sequencing of on-demand ridesharing services. Their analysis indicated on-demand taxi services to improve the service rate, passenger waiting time, and distance traveled per vehicle if a trip delay of five or more minutes is allowed. Bertismas et al. (2018) proposed a tractable algorithm for large-scale real-world vehicle routing. Their algorithm combines the advantages of local search as well as global optimization and yields solutions that are near-optimal.

Another popular on-demand mobility service in urban areas is “bike-sharing”, which allows commuters to hire bicycles for commuting. Generally, shareable bikes are stationed at docks near high-volume pedestrian areas, which have a couple of benefits for cities with traffic problems (Koska et al., 2016). The main advantage of availing bike services is that they do not pollute the environment while at the same time may act as the first- and last-mile commutes for other travel options, such as subways and ferries (DeMaio et al., 2009). However, there are certain challenges associated with bike network operations. One main issue is the possibility of shortages and surplus of bicycles in the bike docks. If there exists a deficit of bikes, then customers are unsatisfied, resulting in revenue loss for the company. On the contrary, excess bikes might result in over-crowding of bikes at a station, due to which customers might have inconveniences to park the bike at the destination docks (Raviv and Kolka, 2013). To overcome issues and facilitate better resource allocation, Li et al. (2015) predicted demand levels at bike docking stations using an algorithm that uses both gradient boosting regression trees and clustering techniques.

While the aforementioned on-demand services use road transport, UAM considers a new dimension for on-demand transportation and has the potential to augment these existing services. Specifically, the UAM concept can be categorized into three groups: (i) small, unmanned vehicles (such as drones) primarily used for package delivery, (ii) autonomously operated air metro services that function similar to public transportation services, and (iii) air taxi services for on-demand mobility of individuals or small groups (National Aeronautics and Space Administration (NASA) et al., 2018a). The study of airborne drones (i.e., group (i)) for logistics operation has been widely investigated (Roca-Riu and Menendez, 2019; Gonzalez et al., 2020; Sah et al., 2020; Salama and Srinivas, 2020). Merkert and Bushell (2020) provided a detailed review of the literature on the critical issues in the current use of drones. We observe that several issues pertaining to the implementation of delivery drones illustrated by Merkert and Bushell (2020), such as security, privacy and acceptance concerns, also exist for urban passenger aviation services (Rajendran and Srinivas, 2020). In recent years, logistics companies have been conducting tests with respect to air taxi design prototypes across the world. The different concept vehicles that have been developed can be classified into one of the three categories - quadrotor, side-by-side, and lift + cruise aircraft (National Aeronautics and Space Administration, 2018b). The pilot and feasibility testing of these concept vehicles have been performed in numerous cities around the globe. For example, Uber has begun a preliminary investigation in Dallas and Los Angeles and has plans to venture into international markets such as Dubai, Tokyo, Singapore, London, and Bangalore (Hawkins, 2018). Another company, Kitty Hawk, announced in early 2018 its goal of launching an air taxi service, *Cora*, in New Zealand (Warwick, 2018). Also, self-piloted eVTOL, *Vahana* by Airbus, has completed flight tests in Oregon (Hawkins, 2018). Additionally, companies like Rolls-Royce, Boeing, and Martin Jetpack are planning to make a debut in the air taxi sector as well (Ridden, 2018).

### 2.2 Demand Prediction for Air Taxi Services

As the concept of air taxi services is still in its early stages, research on forecasting the demand for such services is limited. Besides, there exist numerous challenges in estimating the customer demand for air taxi services. Notably, it is difficult to assess people's opinion of air taxis based on factors such as safety, travel distance, and pre-boarding procedures. Besides, other factors such as pricing strategy, increasing telecommuting trend, and lack of GPS trace data further complicates the task of demand prediction (NASA Mobility UAM Market Study, 2018).

To overcome these challenges, recent works have made efforts to gather subjective opinions of potential customers on different aspects of air taxis services such as willingness to fly and willingness to pay. Garrow et al. (2018) conducted focus groups to estimate the demand for air taxi services and identified wealthy households to embrace it first due to better affordability and higher time value. Besides, they also identified three segments with high demand, namely, daily commuters, airport shuttle, and end-to-end city transfers. To understand the customer's willingness to pay for air taxi services, Binder et al. (2018) developed a stated preference survey. Subsequently, Boddupalli (2019) adapted these surveys and collected responses from 2500 individuals in five major US cities. The authors found commute time, ride fare, and presence of congestion along commute route to be the key determinants of air taxi demand.

To estimate the adoption of air taxi services among people, several recent studies have examined the impact of various factors on consumer's willingness to fly. Winter et al. (2020) obtained data from 510 participants and developed valid statistical models to identify significant predictors of willingness to fly in autonomous air taxis. The authors identified six significant predictors that explained over 76% of the variance in willingness to fly - familiarity, value, fun factor, wariness of new technology, fear and happiness. Another notable work in this domain found nationality and the presence of an automatic parachute system to influence the willingness to fly (Ward et al., 2020). Ragbir et al. (2020) further contributed to this research domain by considering key external factors, such as weather, distance, flight time, and geography. The authors identified that customers were willing to fly more in good weather conditions compared to rainy weather and on shorter flights as opposed to longer flights. Besides, many customers preferred to fly over land instead of water.

While most of the prior works used a qualitative approach for demand prediction, Becker et al. (2018) used a gravity model to predict the market demand for air taxis. They identified 28 cities with the potential for high air taxi demand and ranked them using a multi-criteria method. They found New York City to have the highest potential for air taxis, followed by Los Angeles and Dubai. Likewise, Rajendran and Zack (2019) developed an analytical model for approximating air taxi demand using existing trip data.

### 2.3 Contributions to the Literature

We extend and contribute to the literature in the following ways. First, though very few studies have focused on determining the location of air taxi stations in an urban environment (e.g., Rajendran and Zack, 2019) or evaluating the competitiveness of this soaring everyday transportation method against the regular modes of commutes (e.g., Sun et al., 2018), to the best of our knowledge, this study is the first to predict the demand for air taxi services using a machine learning-based approach. Also, several prior studies that estimate the demand for ridesharing services, such as bike-sharing networks, rarely consider weather-related and location-related components. However, Singhvi et al. (2015) proved the significant impact of these factors on demand prediction. Therefore, in this study, we extend the current work by estimating the demand for air taxis by considering weather-related predictors.

## 3. Data

The estimated air taxi data used for this research is obtained from a prior study by Rajendran and Zack (2019). The authors developed an algorithm to determine the potential air taxi demand using the assumptions suggested by Holden and Goel (2016) and Holden et al. (2018) that (i) all air taxi ride requests are satisfied on an on-demand basis, (ii) all passengers will take only one VTOL leg (i.e., layovers are not considered), and (iii) an individual becomes qualified for an air taxi ride only if the time savings (compared to the ground transportation) is at least 40%.

### 3.1 Data Preparation and Integration

The estimated air taxi record includes the following fields for each trip,

- pickup date and time
- drop-off date and time
- number of passengers transported
- latitude and longitude of origin location
- latitude and longitude of destination location

The predictors and outcome variable required for developing the machine learning model are prepared by aggregating the fields present in the raw trip data. First, to enhance the usability and generalizability of the prediction models, the pickup coordinates (latitude and longitude) are partitioned into different regions (also referred to as location ID) using a clustering algorithm. Specifically, the *k*-means clustering algorithm was chosen for this research as it provides the flexibility to control the number of clusters (i.e., the number of regions) to establish (Krishna and Murty, 1999).

The *k*-means technique is typically used to group the data under consideration into $k$ clusters, where the objective is to minimize the average squared Euclidean distance between the center of a cluster and the data points associated with that cluster. If $\vec{x}$ represents a vector of *V* variables to be clustered (e.g., latitude and longitude) and *K* denotes the total number of clusters to establish, then the procedure for *k*-means clustering is as shown in Algorithm 1, where $\vec{\mu}_k$ represents the centroid of cluster $k$ and $c_k$ denotes the set of elements belonging to cluster $k$ (adapted from Schütze et al., 2008).

Upon clustering, each pickup coordinate is assigned to a specific region or location ID (i.e., cluster centroid). In addition, the pickup date is used to derive ride-related temporal factors - month, day of week and weekday/weekend indicator. Subsequently, the temporal factors are grouped together by location ID, date, and hour of the day.

**Algorithm 1:** Pseudocode for *k*-means Algorithm (adapted from Schütze et al., 2008)

Inputs: $\vec{x} = \{x_1, x_2, \ldots, x_V\}$ and $K$
Initialize $K$ random centroids $(\vec{s}_1, \vec{s}_2, \ldots, \vec{s}_K)$
**for** $k \leftarrow 1$ **to** $K$
**do** $\vec{\mu}_k \leftarrow \vec{s}_k$
**while** stopping criterion has not been met
**do for** $k \leftarrow 1$ **to** $K$
    **do** $c_k \leftarrow \{\}$
    **for** $v \leftarrow 1$ **to** $V$
    **do** $j \leftarrow \arg\min_{j'} |\vec{\mu}_{j'} - \vec{x}_v|$
        $c_j \leftarrow c_j \cup \{\vec{x}_n\}$(reassignment of vectors)
    **for** $k \leftarrow 1$ **to** $K$
    **do** $\vec{\mu}_k \leftarrow \frac{1}{|c_k|} \sum_{\vec{x} \in c_k} \vec{x}$ (recomputation of centroids)
**return** $\{\vec{\mu}_1, \vec{\mu}_2, \ldots, \vec{\mu}_K\}$

In addition to these predictors, weather-related (or environmental) information, namely, temperature, weather condition (snow, rain, normal, etc.), visibility, fog, wind speed, and humidity, are extracted for each hour of day and each date in the dataset using the API of a commercial weather service provider. The output variable is the demand levels (i.e., whether the demand for air taxi is ‘low’, ‘moderate/medium’ or ‘high’), and is obtained by aggregating and binning the number of passengers transported for each location ID, date, and hour of day. A unified data frame is obtained by combining the predictors from two disparate data sources (estimated air taxi ride records and weather-related information) along with the output variable. A summary of the characteristics of the input features and the outcome variable is given in Table 1.

**Table 1:** Description of Data

| Category | Features | Description | Variable Type | Data Source |
|---|---|---|---|---|
| Ride-related | Month | Month of the year during which pickup was made | Categorical (Jan, Feb,…, Dec) | Derived from the data provided by Rajendran and Zack (2019) |
| | Day of week (DOW) | Day of pickup | Categorical (Sun, Mon,…,Sat) | |
| | Time of day (TOD) | Time of day during which customers are picked up (for the sake of conducting analysis, time is categorized into 24 time zones) | Categorical (Time Zone #1, #2,…, #24) | |
| | Weekday Indicator | Whether the day of the ride is a weekday or not | Categorical (Yes or No) | |
| | Location ID | Location ID of the pickup location | Categorical (ID#1, ID#2, ID#3,…, etc.) | Obtained from clustering algorithms |
| Environment-related | Temperature | Temperature at the time of pickup | Continuous | Outside source |
| | Weather condition | Type of weather condition | Categorical (Normal, Snow, Rain & Thunderstorm) | |
| | Visibility | Degree to which an object or light can be clearly noticed | Continuous | |
| | Wind speed | Rate of atmospheric air movement | Continuous | |
| | Humidity | Amount of moisture in the atmosphere | Continuous | |
| | Fog | Whether there is a presence of a thick cloudlike mass or not | Categorical (Yes, No) | |
| Output | Demand Level | Density of passengers requesting air taxi ride at a specific time and location | Categorical (low, moderate, high) | Derived from the data provided by Rajendran and Zack (2019) |

### 3.3 Data Pre-Processing

The unified dataset contained inconsistencies, errors, and missing values, and is typical for any real-world cases. In particular, we observe several discrepancies in the data with zero, negative or irrational values in trip duration and travel distance. Likewise, in some cases, the visibility was recorded as -9999. As a result, the dataset is cleaned or pre-processed before being inputted into the machine learning model. Since it is not possible to retrieve the actual value of these erroneous and missing data, two methods were explored to handle them,

- *imputation* (replacing missing and inconsistent data points with substitute values)
- *listwise deletion/complete case analysis* (removing all rows containing one or more missing values) (Srinivas and Rajendran, 2017).

In the case of the imputation, the following approaches are considered (i) median/mode imputation: substituting the inconsistent data in a continuous predictor with the median value of that predictor and replacing inconsistent data in a categorical feature with the most frequent value within that column and (ii) regression imputation: predicting the incorrect/missing value by regressing it on the other features (Srinivas and Salah, 2020).

To facilitate the training and testing of the machine learning model, each of the levels associated with a categorical variable is converted to a binary feature (0 = No, 1 = Yes) using one-hot encoding technique. For instance, the variable "Month" has 12 levels, and each level will be encoded as an independent binary variable (e.g., encoded feature "Month_Jan" = 1 if the trip is made in January and 0 otherwise). On the other hand, the continuous variables are centered and scaled.

## 4. Demand Prediction using Machine Learning Algorithms

In this research, we develop supervised classification algorithms to predict the demand class using the processed dataset as the output variable is categorical. The objective of a supervised machine learning algorithm is to infer the function (*f*) that maps the relationship between *N* predictors, $\mathbf{x} = \{x_1, x_2, \dots, x_N\}$ and a categorical outcome vector (*y*) based on *M* training examples (historical cases or experiences). To ensure robust model building and evaluation, the processed data is partitioned into training and testing sets. The training dataset feeds both the predictors and corresponding output variable to the machine learning model so that the model can learn/uncover the relationship between them.

To avoid overfitting and estimate the generalization error of the model, a *k*-fold cross-validation resampling procedure is adopted during the learning phase, where the training sample is partitioned into *k* equal-sized subsamples in which *k*-1 subsets are used for fitting the model, and the single retained subset is used for validation (Srinivas and Ravindran, 2018). This procedure is repeated to allow each of the *k* subsets to be used exactly once for validation. Once the model is trained, its classification performance is evaluated using a new unseen dataset (i.e., testing data). However, it is not possible to choose the most appropriate machine learning algorithm (MLA) for a given dataset without any prior knowledge of the nature of the relationship between the predictors and outcome. Therefore, prior studies suggest evaluating multiple MLAs, where the chosen algorithms vary in terms of their learning mechanism/training procedure and computational complexity (Srinivas, 2020). As a result, four popular supervised machine learning algorithms (MLAs) are employed in this study - multinomial logistic regressions, artificial neural networks, random forests and gradient boosting. These four algorithms are chosen because they have demonstrated superior performance for similar classification problems in the literature. A brief description of the four MLAs is provided in the following subsections.

### 4.1 Multinomial Logistic Regression

The multinomial logistic regression model uses the softmax function and a weighted linear combination of the predictors to estimate the probability of each output class, as represented in Figure (2a). During the learning phase, the model infers the weights for each predictor with respect to each class, $\mathrm{b(c)} = \{b_1(c), b_2(c), \dots, b_N(c)\}$, to best map the relationship between the predictors and outcome in the training dataset (Bayaga, 2010). Upon training, the model can estimate the probability of the demand belonging to class $c$ (i.e., low, moderate or high) given an input vector $\vec{x}$, $\mathrm{P}(\mathrm{y} = \mathrm{c}|\vec{x})$, as shown in Equation (1).

$$\mathrm{P}(y = c|\vec{x}) = \frac{e^{(\mathrm{b(c)}\cdot\vec{x})}}{\sum_c e^{(\mathrm{b(c)}\cdot\vec{x})}} \tag{1}$$

### 4.2 Artificial Neural Network

Artificial Neural Network (ANN) is a supervised algorithm, whose learning mechanism is inspired by biological neural networks (Van Gerven and Bohte, 2017). The ANN consists of nodes that behave very similar to neurons of a brain and is represented by three types of layers – input, hidden and output, as shown in Figure (2b). A connection weight is used to link the nodes in two consecutive layers. The input layer represents the set of predictors or features. Each input is multiplied by a weight and is transmitted to the nodes in the hidden layer. While receiving, the hidden layers combine these weighted inputs and relay a function of it to the next layer. Finally, the output layer estimates the probability of each class by using its connection weight and values received from the hidden layer. During ANN training, the key goal is to optimize all the weights to best represent the relationship between the input and output. The backpropagation algorithm is used to update the weights at every iteration of learning by using the gradients of the error function with respect to the connection weight and a learning rate (Van Gerven and Bohte, 2017). If the predicted output is the same as the target output, then the weights are simply reinforced without altering, whereas weights are updated if the prediction is incorrect. Unlike multinomial logistic regression, ANN can uncover complex non-linear relationships between the predictors and output. Nevertheless, the time required to optimize the weights and train the algorithm is expected to be substantially higher than multinomial logistic regression. Upon establishing the weights in the training phase, the ANN can leverage it to predict the demand level for unseen inputs in the testing data.

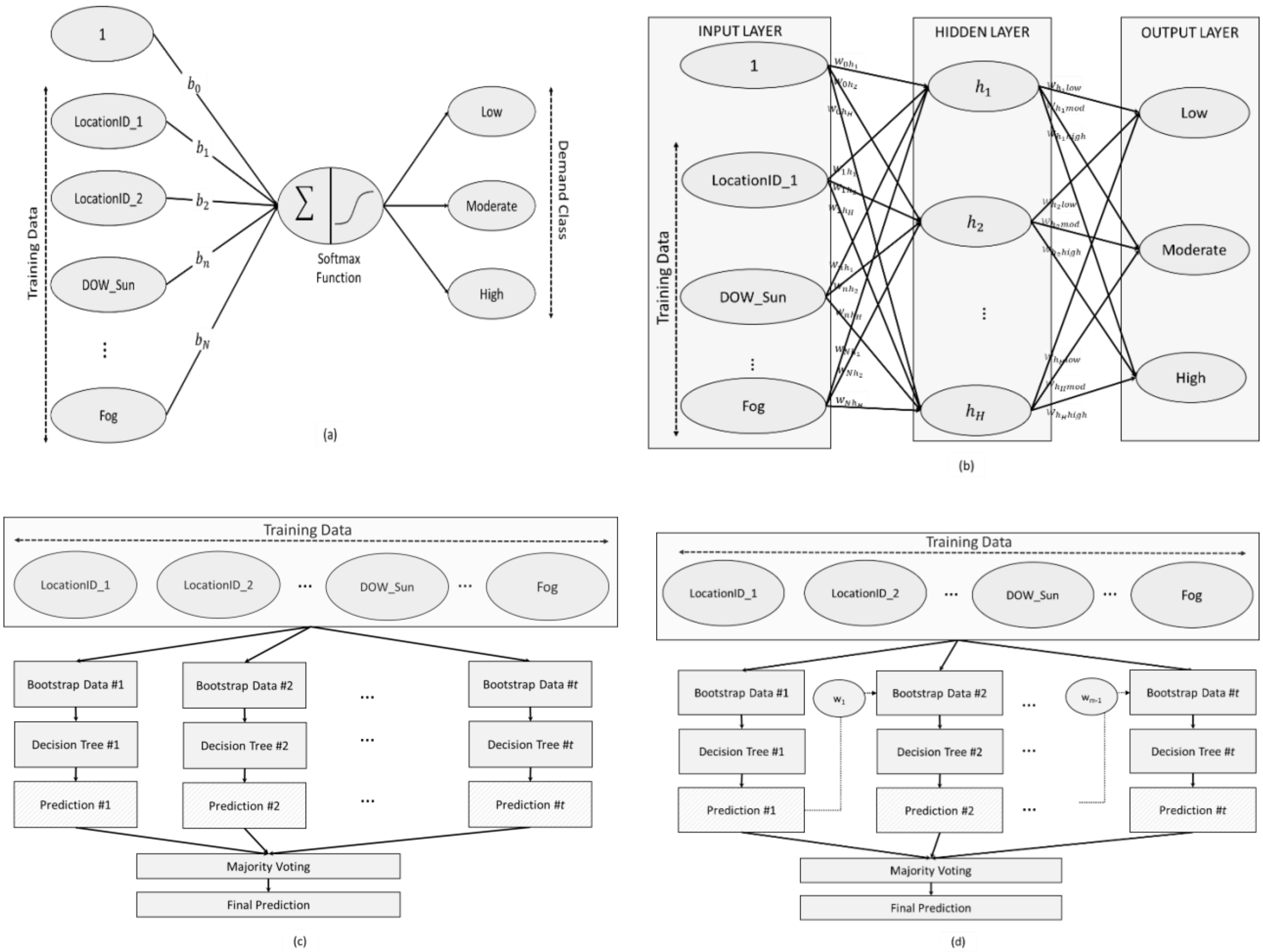

**Figure 2:** Machine Learning Algorithms Evaluated for Predicting Air Taxi Demand - (a) Multinomial Logistic Regression, (b) Artificial Neural Network, (c) Random Forests, and (d) Gradient Boosting

### 4.3 Random Forests

Random forests (RF) is an ensemble algorithm, where several decision trees are trained in parallel to estimate the outcome class (Sarica et al., 2017). The decision trees algorithm has a top-down tree-like structure, where a chosen input variable is recurrently split into two or more categories until a stopping criterion is reached. Typically, the variable that provides the best split at each stage is determined using metrics that measure the effectiveness of a split, such as entropy or information gain. The algorithm usually terminates when there is no substantial improvement in the metrics or when tree depth reaches a certain threshold.

In the case of RF training, a bootstrapped sample of the training data is used to build each decision train (Figure (2c)). Besides, a random subset of predictors is considered to determine the best variable for splitting at a given node of a decision tree. The final class prediction for RF is obtained by taking the majority voting of the predictions obtained from the decision trees. A trained RF algorithm can then predict the demand class for each instance in the testing data. Similar to ANN, RF is capable of uncovering non-linear relationships. However, they adopt a different learning approach when compared to ANN.

### 4.4. Gradient Boosting

Gradient Boosting (GB) algorithm is similar to RF, with regards to the ensemble approach (Biau et al., 2019). However, the decision trees in the GB method are trained sequentially, as illustrated in Figure (2d), and data required for learning is sampled randomly without replacement. Specifically, the "boosting" technique trains several shallow decision trees in series, where each decision tree seeks to rectify the residual error of its predecessors. In other words, once a decision tree is trained, the gradient of the training loss is used to adjust the weights of the next iteration, where the observations that were challenging to classify are given slightly more weights, while those that were easily classified are down-weighted. The trees are iteratively trained with the goal of minimizing a loss function (e.g., negative log likelihood). The final prediction of the demand class is the weighted voting of individual decision trees.

### 4.5 Evaluating Classification Performance of Machine Learning Algorithms

We evaluate the classification performance of the MLAs using three commonly used core evaluation measures for multi-class classification problems, Precision, Recall and $F_1$ score. Precision for class $c$ ($P_c$) is the ratio of correctly classified instances to the total number of cases that were predicted as class $c$. On the other hand, recall for class $c$ ($R_c$) is the number of instances that are correctly classified as class $c$ divided by the total number of cases that actually belong to class $c$. The per-class $F_1$ score is the harmonic mean of precision and recall, as shown in Equation (2).

$$F_1 \text{ score for class c} = 2 \times \frac{P_c \times R_c}{P_c + R_c} \tag{2}$$

The three measures range between 0 and 1, where a higher value indicates better classification performance.

## 5. Results

The study procedure discussed in Section 4 is implemented on R statistical computing software. Besides, a computer configured with Intel i7 quad-core processor, 64 GB RAM, and Windows 10 operating system is used to execute the study procedure. The raw data contained over 3.5 million air taxi eligible trips. A histogram of the estimated average air taxi demand over different days of week and months of a year is illustrated in Figures 3(a) and 3(b), respectively. It can be observed that the estimated air taxi demand varies considerably over different days of the week. In particular, it is at its lowest during the weekend, which could be due to fewer work-related trips. Likewise, the passenger traffic also varies substantially for different months. It is lower at the beginning of the year (January and February), peaks in May, and higher than average during June and the last quarter of the year.

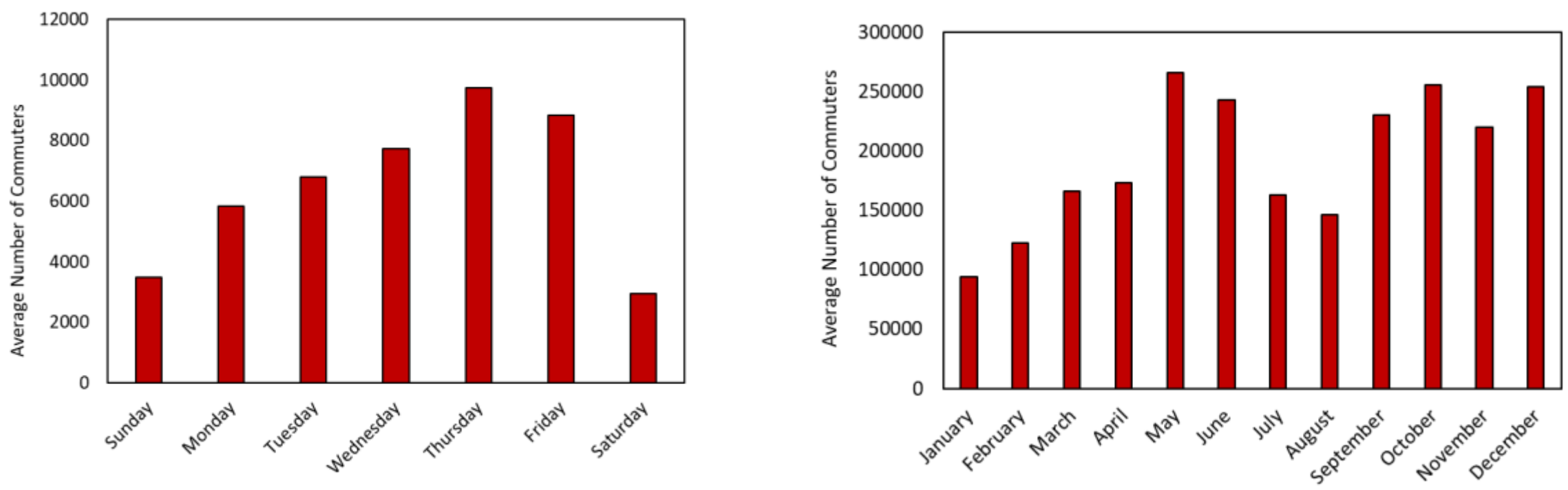

**Figure 3:** Estimated air taxi demand variation over different (a) days of week and (b) months

To prepare the data for machine learning, the pickup coordinates in the raw data is grouped using *k*-means clustering algorithm. As the number of regions established could affect the model performance, four different values of *k* are considered, namely, 5, 10, 15 and 20. Subsequently, the raw data is grouped by the regions, date and day of week, resulting in a consolidated data with more than 200,000 samples on average. Finally, the ride-related temporal- and weather-related factors are derived for these samples, as described in Section 3.1.

On evaluating the three different approaches (mean/mode imputation, regression imputation and listwise deletion) for dealing with inconsistent and missing values, it is observed that the classification performance (i.e., value of the $F_1$ score) did not significantly change between the three techniques. As a result, it is expected that eliminating the missing data is not likely to bias the results significantly. Therefore, the listwise deletion method was chosen for further processing. In addition, we also compared the characteristics of the data retained and data removed. Our results show no substantial difference between the two groups, indicating that the values were missing randomly.

### 5.1 Machine Learning Model Parameters

The MLAs are trained using 70% of the randomly sampled pre-processed records and tested using the remaining 30%. Consistent with prior literature, a 10-fold cross-validation procedure is used during training. LR has no parameters, but the other three algorithms, RF, ANN, and GB are tuned using a grid search method, where an exhaustive search of a specified parameter space is conducted to establish the best set of parameters.

For both the RF and the GB methods, the number of trees is varied from 100 to 1000 in increments of 100. In the case of RF, the values considered for the number of randomly sampled predictors at each decision tree node are $\sqrt{N}$, $N/2$, $N/3$ and $N/4$, where $N$ indicates the total number of predictors (Srinivas and Salah, 2020). For ANN, a feed-forward algorithm is developed with three layers (input, hidden, and output) coupled with backpropagation learning. While the input and output layers are known in advance based on the independent and dependent variables, the number of nodes in the hidden layer is varied between 1 and $N$, in increments of 5. Furthermore, 0.01, 0.05, and 0.10 learning rate values are investigated for training the ANN.

For all the four region partitioning settings, the classification performance of RF and GB algorithms did not improve significantly beyond 300 trees. Besides, the best setting for the number of predictors to be randomly sampled at each decision tree node is established to be $\sqrt{N}$. In the case of ANN, the best value for the learning rate and the number of nodes in the hidden layer is found to be 0.05 and 30, respectively.

**5.2 Predictive Performance of Machine Learning Algorithms**

Table 2 presents the classification performance of the four MLAs on the testing dataset. The LR model consistently underperformed when compared to the other three MLAs, irrespective of the number of regions (or stations) considered. On the other hand, with regards to the $F_1$ score, the GB algorithm consistently emerged as the best method for classifying the demand levels. Moreover, GB and RF algorithms invariably produced higher per-class precision, recall and $F_1$ score for low and high demand levels as opposed to moderate demand levels, whereas the ANN algorithm achieved the highest per-class performance for classifying moderate demand level. The pattern exhibited in the classification performance of the MLAs for each class could be attributed to their learning mechanism. In other words, the learning approach adopted by RF and GB appears to accurately map the relationship between the predictors for low as well high demand levels, while ANN's learning process better fits the data for moderate demand levels. In summary, it can be observed that MLAs are capable of estimating the demand level with considerable accuracy.

The computational time needed to train each machine learning algorithm is shown in Table 3. For the data under study, LR took the least amount of time for training, which is several orders of magnitude smaller than the other methods. This is not surprising, considering the learning process of LR. The other three learning methods are similar in terms of orders of magnitude and do not seem to be noticeably different. GB has the second least computational time in regard to training and is followed by ANN and R.

**Table 2:** Performance of different machine learning algorithms for air taxi demand level prediction

| No. of Clusters | Demand | Logistic Regression | | | ANN | | | Random Forest | | | Gradient Boosting | | |
|---|---|---|---|---|---|---|---|---|---|---|---|---|---|
| | | Precision | Recall | F1-Score | Precision | Recall | F1-Score | Precision | Recall | F1-Score | Precision | Recall | F1-Score |
| K=5 | Low | 0.5032 | 0.3719 | 0.4277 | 0.5195 | 0.5593 | 0.5387 | 0.8857 | 0.6500 | 0.7497 | 0.8999 | 0.7215 | 0.8009 |
| | Medium | 0.4618 | 0.3293 | 0.3844 | 0.6294 | 0.5412 | 0.5820 | 0.3854 | 0.5282 | 0.4457 | 0.4453 | 0.5009 | 0.4715 |
| | High | 0.3593 | 0.4902 | 0.4147 | 0.4201 | 0.5414 | 0.4731 | 0.6721 | 0.6374 | 0.6543 | 0.6912 | 0.6987 | 0.6949 |
| | Average | 0.4414 | 0.3971 | 0.4089 | 0.5230 | 0.5473 | 0.5312 | 0.6477 | 0.6052 | 0.6165 | 0.6788 | 0.6404 | 0.6558 |
| K=10 | Low | 0.5035 | 0.3680 | 0.4252 | 0.5793 | 0.4947 | 0.5337 | 0.8179 | 0.6657 | 0.7340 | 0.7767 | 0.7138 | 0.7439 |
| | Medium | 0.4627 | 0.3344 | 0.3882 | 0.5467 | 0.4632 | 0.5015 | 0.3547 | 0.4229 | 0.3858 | 0.3707 | 0.4509 | 0.4069 |
| | High | 0.3585 | 0.4838 | 0.4118 | 0.4399 | 0.5813 | 0.5008 | 0.8425 | 0.6250 | 0.7177 | 0.8759 | 0.6797 | 0.7654 |
| | Average | 0.4416 | 0.3954 | 0.4084 | 0.5220 | 0.5131 | 0.5120 | 0.6717 | 0.5712 | 0.6125 | 0.6744 | 0.6148 | 0.6387 |
| K=15 | Low | 0.5152 | 0.5083 | 0.5118 | 0.5813 | 0.5818 | 0.5815 | 0.7824 | 0.7326 | 0.7567 | 0.8143 | 0.7224 | 0.7656 |
| | Medium | 0.4087 | 0.3560 | 0.3805 | 0.5173 | 0.4430 | 0.4772 | 0.3520 | 0.4253 | 0.3852 | 0.3700 | 0.4351 | 0.3999 |
| | High | 0.3615 | 0.3367 | 0.3487 | 0.4890 | 0.3655 | 0.4184 | 0.8399 | 0.7010 | 0.7642 | 0.8855 | 0.6529 | 0.7516 |
| | Average | 0.4285 | 0.4003 | 0.4136 | 0.5292 | 0.4634 | 0.4924 | 0.6581 | 0.6196 | 0.6354 | 0.6899 | 0.6035 | 0.6390 |
| K=20 | Low | 0.4773 | 0.4677 | 0.4725 | 0.5002 | 0.6018 | 0.5463 | 0.8039 | 0.7480 | 0.7749 | 0.8223 | 0.8124 | 0.8173 |
| | Medium | 0.3326 | 0.3332 | 0.3329 | 0.5661 | 0.5756 | 0.5708 | 0.3478 | 0.3898 | 0.3676 | 0.3944 | 0.4389 | 0.4155 |
| | High | 0.3928 | 0.3865 | 0.3896 | 0.4607 | 0.4034 | 0.4302 | 0.9274 | 0.7133 | 0.8064 | 0.8580 | 0.6011 | 0.7069 |
| | Average | 0.4009 | 0.3958 | 0.3983 | 0.5090 | 0.5269 | 0.5158 | 0.6930 | 0.6170 | 0.6496 | 0.6916 | 0.6175 | 0.6466 |

**Table 3:** Computational Time (in seconds)

| # of stations | Logistic Regression | Artificial Neural Network | Random Forest | Gradient Boosting |
|---|---|---|---|---|
| 5 | 1.87 | 902.48 | 940.49 | 740.00 |
| 10 | 3.01 | 1075.45 | 1229.45 | 990.10 |
| 15 | 4.38 | 1531.75 | 1583.47 | 1310.31 |
| 20 | 5.27 | 1754.85 | 2186.01 | 1753.02 |

Table 4 shows the important features for predicting the demand level using GB algorithm. We specifically focus on this method due to its superior classification performance across different settings (inferred from Table 2). The feature importance is obtained by randomly permuting its value and assessing the corresponding impact on the classification error. The station located near JFK International Airport (LocationID_1), weekday indicator, and time periods 4 – 6 PM (i.e., TimeSlot_16 – TimeSlot_18) consistently emerge as the top predictors in all settings. Specifically, "LocationID_1" appears to be the most important predictor followed by "Weekday Indicator" for all settings except for the case with 20 regions/stations, where the ranks are reversed. Besides, the importance of a variable in predicting the outcome is observed to change for different station settings. For instance, when the number of stations is restricted to a small number (i.e., say 5), we observe timeslots 9 – 11 PM and weather conditions (e.g., presence of rain) to be among the top features, but does not exhibit the same importance when the stations established are increased to 10, 15 or 20. This could be due to the change in the characteristics of the data when increasing the number of stations. Specifically, when the number of stations is increased (say from 5 to 10), the subset of data (or rows) pertaining to a specific station can also change. For example, LocationID_1 may have 5000 rows associated with it if the total number of locations is 5 but decrease to 3000 rows when the number of locations is increased to 10 (as the remaining 2000 rows may be assigned to LocationID_6 – LocationID_10). Thus, depending on the number of stations, the order of importance of certain predictors are changed, while some are entirely different.

**Table 4:** Top Features using GBM for Different Stations

| # of Stations | Top Features using GBM | | | | |
|---|---|---|---|---|---|
| | #1 | #2 | #3 | #4 | #5 |
| 5 | Location ID_1 (near JFK) | Weekday | Timeslot_16 - Timeslot_18 (4 - 6 PM) | Timeslot_21 - Timeslot_23 (9 - 11 PM) | Weather_Rain (presence of rain) |
| 10 | Location ID_1 (near JFK) | Weekday | Timeslot_16 - Timeslot_18 (4 - 6 PM) | Month_Jan (January) | Temperature |
| 15 | Location ID_1 (near JFK) | Weekday | Timeslot_16 - Timeslot_18 (4 - 6 PM) | LocationID_5 (intersection of Carlton Avenue and Boerum Street in Brooklyn) | LocationID_7 (Sedgwick Ave, The Bronx) |
| 20 | Weekday | Location ID_1 (near JFK) | Temperature | Timeslot_16 - Timeslot_18 (4 - 6 PM) | LocationID_5 (intersection of Carlton Avenue and Boerum Street in Brooklyn) |

**6. Discussion**

This research considers the use of machine learning approaches for predicting the air taxi demand level across different time periods and locations. Besides, it adds to the existing study of Rajendran and Zack (2019) by partitioning demand points into several regions and estimating the demand level for each region as opposed to only determining the potential air taxi stations. The results clearly demonstrate the suitability of MLAs in estimating the demand levels and have many practical implications. First, the predictions from the MLA can be leveraged by route optimization algorithms to dispatch and schedule air taxis to different regions effectively. Such an approach would enable logistics companies to achieve a balance between customer waiting time and vehicle idle time. Second, the demand estimates can be aggregated and used for determining the number of takeoff/landing pad in a skyport. Likewise, the aggregated forecasts can aid in purchasing decisions, such as fleet procurement. Finally, the demand projections obtained from this research can also be capitalized for establishing the pricing strategies for air taxi services, especially the dynamic pricing structure where the ride fare is altered in real-time (Rajendran and Srinivas, 2020).

This paper also identifies the most prominent features for estimating the demand level. Consistent with prior studies, we found weather-related features, such as temperature and presence of rain, to be important variables factors for predicting the spatiotemporal demand (Yao et al., 2018; Kim et al., 2020; Liu et al., 2020). Earlier studies also uncovered the relationship between time of day and regular taxi demand (Yang and Gonzales, 2014), which is also concurred with our findings. While most of the top features (such as JFK, Weekday Indicator, and 4 - 6 PM) are not significantly impacted by the number of stations, certain variables (such as 9 - 11 PM, presence of rain, temperature, and Carlton Ave, Brooklyn) vary based on the spatial factor. The insights on important predictors could help the decision-makers in strategic choices, such as the number of vertiports/vertistops to establish and its location.

Despite the merits and implications of this study, there are a few notable limitations. First, the predictors considered in this research is limited by the air taxi trip estimates provided in a prior study by Rajendran and Zack (2019). Nevertheless, with the emergence of the latent air taxi demand data, other potential predictors, such as ride fare, can be incorporated, and the study procedure presented in this paper can be adapted due its generic nature. Second, the performance of the MLAs is investigated only using the data corresponding to one metropolitan city in the US. To generalize the findings of this study, future work could consider predicting spatiotemporal air taxi demand levels in other metropolitan cities. Finally, the MLAs considered in this research are limited to four well-performing algorithms in the literature. Potential future scope for improving the prediction performance would be to consider other MLAs, such as deep neural networks as well as stacking algorithms (which combines predictions from multiple MLAs using base- and meta-learners).

## 7. Conclusions

With the rapid expansion of air transportation services, research on urban air mobility (UAM), in particular, air taxi, is being widely investigated by academicians and practitioners. This study is one of the first to predict the demand level (low, moderate, high) for air taxis across different time periods and locations. Several ride-related (e.g., pickup and drop-off locations, distance, time of the day, day of the week) and environment-related (e.g., temperature, presence of rain or snow, visibility) factors are considered as predictors. We develop four machine learning models, namely, multinomial logistic regression, artificial neural network, random forests and gradient boosting method to map the relationship between the predictors and demand level. The performance of these models is evaluated using three metrics - precision, recall, and $F_1$ score. The results demonstrate the capability of machine learning models to predict the demand class and also show gradient boosting algorithm to consistently achieve the best classification performance. Besides, our study also revealed several critical predictors for estimating the demand levels. The insights obtained from the study have numerous practical implications and aid decision-makers in strategic, tactical, and operational decisions.

**References**

Agatz, N., Erera, A. L., Savelsbergh, M. W., & Wang, X. (2011). Dynamic ride-sharing: A simulation study in metro Atlanta. Procedia-Social and Behavioral Sciences, 17, 532-550.

Agatz, N., Erera, A., Savelsbergh, M., & Wang, X. (2012). Optimization for dynamic ride-sharing: A review. European Journal of Operational Research, 223(2), 295-303.

Alonso-Mora, J., Samaranayake, S., Wallar, A., Frazzoli, E., & Rus, D. (2017). On-demand high-capacity ride-sharing via dynamic trip-vehicle assignment. Proceedings of the National Academy of Sciences, 114(3), 462-467.

Bayaga, A. (2010). Multinomial Logistic Regression: Usage and Application in Risk Analysis. Journal of applied quantitative methods, 5(2).

Biau, G., Cadre, B., & Rouvìère, L. (2019). Accelerated gradient boosting. Machine Learning, 108(6), 971-992.

Chang, H. W., Tai, Y. C., & Hsu, J. Y. J. (2010). Context-aware taxi demand hotspots prediction. International Journal of Business Intelligence and Data Mining, 5(1), 3-18.

Chen, P. Y., Liu, J. W., & Chen, W. T. (2010, September). A fuel-saving and pollution-reducing dynamic taxi-sharing protocol in VANETs. In Vehicular Technology Conference Fall (VTC 2010-Fall), 2010 IEEE 72nd (pp. 1-5). IEEE.

DeMaio, P. (2009). Bike-sharing: History, impacts, models of provision, and future. Journal of public transportation, 12(4), 3.

Garrett-Glaser, Brian (2019). https://www.aviationtoday.com/2019/11/26/world-economic-forum-city-los-angeles-launch-working-group-urban-air-mobility/

Gonzalez-R, P. L., Canca, D., Andrade-Pineda, J. L., Calle, M., & Leon-Blanco, J. M. (2020). Truck-drone team logistics: A heuristic approach to multi-drop route planning. Transportation Research Part C: Emerging Technologies, 114, 657-680.

Gurumurthy, K. M., & Kockelman, K. M. (2018). Analyzing the dynamic ride-sharing potential for shared autonomous vehicle fleets using cellphone data from Orlando, Florida. Computers, Environment and Urban Systems, 71, 177-185.

Hasan, S. (2019). Urban Air Mobility (UAM) Market Study.

Hawkins, Andrew J. "Airbus' Autonomous 'Air Taxi' Vahana Completes Its First Test Flight." The Verge, The Verge, 1 Feb. 2018, www.theverge.com/2018/2/1/16961688/airbus-vahana-evtol-first-test-flight

Holden and Goel. "Fast-Forwarding to a Future of On-Demand Urban Air Transportation." 27 October 2016

Holden, J., Allison, E., Goel, N., and Swaintek, S, (2018). Session presented at the meeting of the Uber Keynote: Scaling UberAir.

Holguín-Veras, J., Ozbay, K., Kornhauser, A., Brom, M. A., Iyer, S., Yushimito, W. F., ... & Silas, M. A. (2011). Overall impacts of off-hour delivery programs in New York City Metropolitan Area. Transportation Research Record, 2238(1), 68-76.

Jeffrey-Wilensky, Jaclyn. “Uber, Lyft Say They Help Ease Traffic Congestion. New Study Says Otherwise.” NBCNews.com, NBCUniversal News Group, 8 May 2019, www.nbcnews.com/mach/science/ride-sharing-firms-say-they-help-ease-traffic-congestion-new-ncna1003051.

Johnson, W., & Silva, C. (2018). Observations from Exploration of VTOL Urban Air Mobility Designs.

Jung, J., Jayakrishnan, R., & Park, J. Y. (2013). Design and modeling of real-time shared-taxi dispatch algorithms. In Proc. Transportation Research Board 92nd Annual Meeting.

Kim, T., Sharda, S., Zhou, X., & Pendyala, R. M. (2020). A stepwise interpretable machine learning framework using linear regression (LR) and long short-term memory (LSTM): City-wide demand-side prediction of yellow taxi and for-hire vehicle (FHV) service. Transportation Research Part C: Emerging Technologies, 120, 102786.

Koska, T., & Rudolph, F. (2016). The Role of Walking and Cycling in Reducing Congestion: A Portfolio of Measures. Brussels. Available at http://www.h2020-flow.eu.

Krishna, K., & Murty, M. N. (1999). Genetic K-means algorithm. IEEE Transactions on Systems, Man, and Cybernetics, Part B (Cybernetics), 29(3), 433-439.

Li, Y., Lu, J., Zhang, L., & Zhao, Y. (2017). Taxi booking mobile app order demand prediction based on short-term traffic forecasting. Transportation Research Record, 2634(1), 57-68.

Li, Y., Zheng, Y., Zhang, H., & Chen, L. (2015). Traffic prediction in a bike-sharing system. In Proceedings of the 23rd SIGSPATIAL International Conference on Advances in Geographic Information Systems (p. 33). ACM.

Liao, S., Zhou, L., Di, X., Yuan, B., & Xiong, J. (2018, January). Large-scale short-term urban taxi demand forecasting using deep learning. In 2018 23rd Asia and South Pacific Design Automation Conference (ASP-DAC) (pp. 428-433). IEEE.

Lin, Y., Li, W., Qiu, F., & Xu, H. (2012). Research on optimization of vehicle routing problem for ride-sharing taxi. Procedia-Social and Behavioral Sciences, 43, 494-502.

Liu, Y., Liu, Z., Lyu, C., & Ye, J. (2019). Attention-based deep ensemble net for large-scale online taxi-hailing demand prediction. IEEE Transactions on Intelligent Transportation Systems.

Liu, Z., Chen, H., Li, Y., & Zhang, Q. (2020). Taxi demand prediction based on a combination forecasting model in hotspots. Journal of Advanced Transportation, 2020.

Lokhandwala, M., & Cai, H. (2018). Dynamic ride sharing using traditional taxis and shared autonomous taxis: A case study of NYC. Transportation Research Part C: Emerging Technologies, 97, 45-60.

Ma, S., Zheng, Y., & Wolfson, O. (2013, April). T-share: A large-scale dynamic taxi ridesharing service. In Data Engineering (ICDE), 2013 IEEE 29th International Conference on (pp. 410-421). IEEE.

Ma, S., Zheng, Y., & Wolfson, O. (2015). Real-Time City-Scale Taxi Ridesharing. IEEE Trans. Knowl. Data Eng., 27(7), 1782-1795.

Maciejewski, M., Bischoff, J., & Nagel, K. (2016). An assignment-based approach to efficient real-time city-scale taxi dispatching. IEEE Intelligent Systems, 31(1), 68-77.

Matsumoto, H., & Domae, K. (2018). The effects of new international airports and air-freight integrator's hubs on the mobility of cities in urban hierarchies: A case study in East and Southeast Asia. Journal of Air Transport Management, 71, 160-166.

Merkert, R., & Bushell, J. (2020). Managing the drone revolution: A systematic literature review into the current use of airborne drones and future strategic directions for their effective control. Journal of Air Transport Management, 89, 101929.

Moreira-Matias, L., Gama, J., Ferreira, M., Mendes-Moreira, J., & Damas, L. (2013). Predicting taxi–passenger demand using streaming data. IEEE Transactions on Intelligent Transportation Systems, 14(3), 1393-1402.

National Aeronautics and Space Administration (NASA), 2018b. In: Observations from Exploration ofVTOL Urban Air Mobility Designs. NASA. https://rotorcraft.arc.nasa.gov/Research/Programs/eVTOL_observations_Johnson_Silva_2018.pdf

National Aeronautics and Space Administration (NASA), Crown Consulting, McKinsey and Company, Ascension Global, Georgia Tech Aerospace Systems Design Lab, 2018a. In: Urban Air Mobility (UAM) Market Study. NASA. https://www.nasa.gov/sites/default/files/atoms/files/uam-market-study-executive-summary-v2.pdf

Ota, M., Vo, H., Silva, C., & Freire, J. (2015, October). A scalable approach for data-driven taxi ride-sharing simulation. In Big Data (Big Data), 2015 IEEE International Conference on(pp. 888-897). IEEE.

Partnership for New York City . "Growth or Gridlock: The Economic Case for Traffic Relief and Transit Improvement for a Greater New York." Partnership for New York City, Partnership for New York City , 2019, www.pfnyc.org/reports/GrowthGridlock_4pg.pdf.

Patnoe, L. (2018). Flyshare 2020 (Doctoral dissertation).

Pillac, V., Gendreau, M., Guéret, C., & Medaglia, A. L. (2013). A review of dynamic vehicle routing problems. European Journal of Operational Research, 225(1), 1-11.

Posen, H. A. (2015). Ridesharing in the sharing economy: Should regulators impose Uber regulations on Uber. Iowa L. Rev., 101, 405.

Ragbir, N. K., Rice, S., Winter, S. R., Choy, E. C., & Milner, M. N. (2020). How weather, distance, flight time, and geography affect consumer willingness to fly in autonomous air taxis. The Collegiate Aviation Review International, 38(1).

Rajendran, S & Shulman, J. (2020). Study of Emerging Air Taxi Network Operation using Discrete-Event Systems Simulation Approach. Journal of Air Transport Management (in press).

Rajendran, S., & Pagel, E. (2020). Recommendations for emerging air taxi network operations based on online review analysis of helicopter services. Heliyon, 6(12), e05581.

Srinivas, S. and Rajendran (2017). A Data-Driven Approach for Multiobjective Loan Portfolio Optimization Using Machine-Learning Algorithms and Mathematical Programming. In Big Data Analytics Using Multiple Criteria Decision-Making Models (pp. 191-226). CRC Press.

Rajendran, S., & Zack, J. (2019). Insights on strategic air taxi network infrastructure locations using an iterative constrained clustering approach. Transportation Research Part E: Logistics and Transportation Review, 128, 470-505.

Raviv, T., & Kolka, O. (2013). Optimal inventory management of a bike-sharing station. Iie Transactions, 45(10), 1077-1093.

Roca-Riu, M., & Menendez, M. (2019). Logistic deliveries with drones: State of the art of practice and research. In 19th Swiss Transport Research Conference (STRC 2019). STRC.

Sah, B., Gupta, R., & Bani-Hani, D. (2020). Analysis of barriers to implement drone logistics. International Journal of Logistics Research and Applications, 1-20.

Salama, M., & Srinivas, S. (2020). Joint optimization of customer location clustering and drone-based routing for last-mile deliveries. Transportation Research Part C: Emerging Technologies, 114, 620-642.

Santos, D. O., & Xavier, E. C. (2015). Taxi and ride sharing: A dynamic dial-a-ride problem with money as an incentive. Expert Systems with Applications, 42(19), 6728-6737.

Sarica, A., Cerasa, A., & Quattrone, A. (2017). Random forest algorithm for the classification of neuroimaging data in Alzheimer's disease: a systematic review. Frontiers in aging neuroscience, 9, 329.

Schütze, H., Manning, C. D., & Raghavan, P. (2008). Introduction to information retrieval (Vol. 39, pp. 234-265). Cambridge: Cambridge University Press.

Sebastian, Willy. “New York Taxi Trips Analysis.” New York Taxi Trips Analysis, 2 June 2018, rpubs.com/willyarrows/NYCTaxiTripsAnalysis.

Singhvi, D., Singhvi, S., Frazier, P. I., Henderson, S. G., O'Mahony, E., Shmoys, D. B., & Woodard, D. B. (2015, April). Predicting bike usage for new york city’s bike sharing system. In Workshops at the Twenty-Ninth AAAI Conference on Artificial Intelligence.

Smith, R. D. (1994). Safe Heliports Through Design and Planning. A Summary of FAA Research and Development (No. DOT/FAA/RD-93/17). Federal Aviation Administration Washington DC Systems Research and Development Service.

Srinivas, S. (2020). A machine learning-based approach for predicting patient punctuality in ambulatory care centers. International Journal of Environmental Research and Public Health, 17(10), 3703.

Straubinger, A., Rothfeld, R., Shamiyeh, M., Büchter, K. D., Kaiser, J., & Plötner, K. O. (2020). An overview of current research and developments in urban air mobility–Setting the scene for UAM introduction. Journal of Air Transport Management, 87, 101852.

Sun, X., Wandelt, S., & Stumpf, E. (2018). Competitiveness of on-demand air taxis regarding door-to-door travel time: A race through Europe. Transportation Research Part E: Logistics and Transportation Review, 119, 1-18.

Van Gerven, M., & Bohte, S. (2017). Artificial neural networks as models of neural information processing. Frontiers in Computational Neuroscience, 11, 114.

Ward, K. A., Winter, S. R., Cross, D. S., Robbins, J. M., Mehta, R., Doherty, S., & Rice, S. Safety systems, culture, and willingness to fly in autonomous air taxis: A multi-study and mediation analysis. Journal of Air Transport Management, 91, 101975.

Warwick, G. (2018). New Zealand welcomes flight tests of Kitty Hawk's eVTOL air taxi: full-scale prototypes of Cora air taxi in flight testing; transitional eVTOL combines rotors for vertical flight with wings for efficient forward flight. Aviation Week & Space Technology.

Winter, S. R., Rice, S., & Lamb, T. L. (2020). A prediction model of Consumer' willingness to fly in autonomous air taxis. Journal of Air Transport Management, 89, 101926.

Witt, A., Suzor, N., & Wikström, P. (2015). Regulating ride-sharing in the peer economy. Communication Research and Practice, 1(2), 174-190.

Wong, K. I., & Bell, M. G. (2006). The optimal dispatching of taxis under congestion: A rolling horizon approach. Journal of advanced transportation, 40(2), 203-220.

Yang, C., & Gonzales, E. J. (2014). Modeling taxi trip demand by time of day in New York City. Transportation Research Record, 2429(1), 110-120.

Yao, H., Wu, F., Ke, J., Tang, X., Jia, Y., Lu, S., ... & Li, Z. (2018, April). Deep multi-view spatial-temporal network for taxi demand prediction. In Proceedings of the AAAI Conference on Artificial Intelligence (Vol. 32, No. 1).

Yao, H., Wu, F., Ke, J., Tang, X., Jia, Y., Lu, S., ... & Li, Z. (2018, April). Deep multi-view spatial-temporal network for taxi demand prediction. In Proceedings of the AAAI Conference on Artificial Intelligence (Vol. 32, No. 1).

Zhang, K., Feng, Z., Chen, S., Huang, K., & Wang, G. (2016, June). A framework for passengers demand prediction and recommendation. In 2016 IEEE International Conference on Services Computing (SCC) (pp. 340-347). IEEE.

Zhao, K., Khryashchev, D., Freire, J., Silva, C., & Vo, H. (2016, December). Predicting taxi demand at high spatial resolution: Approaching the limit of predictability. In 2016 ieee international conference on Big data (big data) (pp. 833-842). IEEE.